\documentclass[letterpaper, 10 pt, conference]{ieeeconf}  

\usepackage{graphicx}
\usepackage{float}
\usepackage{amsmath}
\usepackage{threeparttable, tablefootnote}

\IEEEoverridecommandlockouts                              

\floatstyle{ruled}

\usepackage{subfigure}

\let\vec\boldvec

\DeclareMathOperator*{\argmin}{\textbf{argmin}}
\usepackage{algpseudocode}
\usepackage{epstopdf}
\usepackage{multicol}
\usepackage{booktabs}

\usepackage[hidelinks]{hyperref}
\usepackage{amssymb}
\usepackage{bm}
\newcommand{\trsp}{{\!\scriptscriptstyle\top}}
\usepackage{arydshln}

\usepackage{cite}

\newcommand{\myparagraphnp}[1]{
\vspace{0cm}\noindent
\textbf{#1}
}

\usepackage[linesnumbered, ruled]{algorithm2e}
\SetKwRepeat{Do}{do}{while}

\title{\LARGE \bf
EKMP: Generalized Imitation Learning with Adaptation, Nonlinear Hard Constraints and Obstacle Avoidance

}
\author{Yanlong Huang
\thanks{Yanlong Huang is with School of Computing, University of Leeds, Leeds LS29JT, UK. \tt\small y.l.huang@leeds.ac.uk}
}

\begin{document}

\maketitle
\thispagestyle{empty}
\pagestyle{empty}

\begin{abstract}
As  a user-friendly and straightforward solution for robot trajectory generation, imitation learning has been viewed as a vital direction in the context of robot skill learning. In contrast to unconstrained imitation learning which ignores possible internal and external constraints arising from environments and robot kinematics/dynamics, recent works on constrained imitation learning allow for transferring human skills to unstructured scenarios, further enlarging the application domain of imitation learning. While various constraints have been studied, e.g.,  joint limits, obstacle avoidance and plane constraints, the problem of nonlinear hard constraints has not been well-addressed. In this paper, we propose \emph{extended kernelized movement primitives} (EKMP) to cope with most of the key problems in imitation learning, including nonlinear hard constraints. Specifically, EKMP is capable of learning the probabilistic features of multiple demonstrations, adapting the learned skills towards arbitrary desired points in terms of joint position and velocity, avoiding obstacles at the level of robot links, as well as satisfying arbitrary linear and nonlinear, equality and inequality hard constraints. Besides, the connections between EKMP and state-of-the-art motion planning approaches are discussed. Several evaluations including the planning of joint trajectories for a 7-DoF robotic arm are provided to verify the effectiveness of our framework.
\end{abstract}

\section{Introduction}
In a myriad of applications, e.g., grasping \cite{stulp2011learning}, reaching \cite{howard2009novel,ude2010task} and painting \cite{silverio2019uncertainty} tasks,
appropriate planning of robot trajectories in either Cartesian space or joint space is crucial.  In this line, imitation learning becomes appealing due to its intuitive  and efficient way of transferring human skills to robots. In addition to the reproduction of demonstrated skills, imitation learning aims at adapting the learned skills to new situations without any additional demonstrations, which is highly desirable towards reducing human intervention. Many algorithms have provided this adaptation feature, such as dynamical movement primitives (DMP) \cite{ijspeert} and task-parameterized Gaussian mixture model \cite{calinon2016} (GMM),
whereas it is non-trivial to directly employ them to constrained scenarios,
such as obstacle avoidance, equality (e.g., plane constraints in task space) and inequality (e.g., joint limits) constraints, albeit that constraints are almost ubiquitous in various robotics systems.

In order to cope with dynamical environments, many remarkable algorithms have been  developed. For example,  in \cite{park2008movement} potential field and null-space adjustment were introduced into DMP  to avoid collisions of robot 
links with obstacles.  Reinforcement learning was combined with imitation learning in \cite{koert2016demonstration, huang2018generalized} so as to find collision-free trajectories.
Functional gradient was employed in \cite{osa2017guiding, shyam2019improving} to optimize an objective function comprising  imitation learning and obstacle avoidance costs. 
However, none of the aforementioned works \cite{park2008movement, koert2016demonstration, huang2018generalized, osa2017guiding, shyam2019improving} takes hard constraints into account. In \cite{huang2020linearly,saveriano2019learning}, the problem of imitation learning under linear hard constraints was studied, whereas obstacle avoidance and in particular nonlinear constraints were not addressed, hence handicapping their applications to more complicated tasks.

In this paper, we aim for a novel framework capable of addressing most of the key problems in imitation learning. Specifically, the proposed approach can:
\begin{enumerate}
	\item[(i)] \emph{learn} the probabilistic properties of multiple demonstrations and \emph{adapt} the learned skills to new situations;
	\item[(ii)] \emph{avoid collisions} between robot links and obstacles;
	\item[(iii)] satisfy linear and nonlinear \emph{hard constraints} in terms of equality and inequality;
	\item[(iv)] learn joint position and velocity
	  simultaneously and maintain the corresponding \emph{derivative relationship} while avoiding obstacles and meeting hard constraints;
\end{enumerate}

As  kernelized movement primitives (KMP) \cite{huang2017} has exhibited reliable performance
in many tasks, e.g., human robot collaboration \cite{silverio2019uncertainty,huang2017}, walking tasks in humanoid \cite{ding2020robust} and exoskeleton robots \cite{zou2020learning}, we propose to extend KMP towards developing a more generic constrained imitation learning framework, which we will refer to as \emph{extended KMP} (EKMP). More concretely, we explain the rationale of EKMP in Section~\ref{sec:ekmp}. Subsequently, we discuss connections between EKMP and state-of-the-art motion planning algorithms in Section~\ref{sec:connection}, including covariant Hamiltonian optimization
for motion planning (CHOMP) \cite{ratliff}, Gaussian process motion planner (GPMP) \cite{mukadam2016gaussian} and motion planning in reproducing kernel Hilbert space (RKHS)  \cite{marinho2016functional}. 
Finally, we showcase the performance of EKMP through several evaluations 
in Section~\ref{sec:eva}.

\section{Extended Kernelized Movement Primitives \label{sec:ekmp}}
In this section, 
we first briefly describe the probabilistic modeling of multiple demonstrations in Section~\ref{subsec:gmm}. After that, we exploit the extracted probabilistic features
and derive EKMP in Section~\ref{subsec:ekmp:problem} and \ref{subsec:ekmp}, as well as illustrate the implementation of EKMP in Section~\ref{subsec:ekmp:imple}.
Some constraints (e.g., linear constraints) are discussed as special cases of EKMP
in Section~\ref{subsec:ekmp:extend}.

\subsection{Learning Probabilistic Features of Demonstrations \label{subsec:gmm}}
Suppose  we have access to a set of demonstrations in the form of time $t$, joint position $\vec{q}\in \mathbb{R}^\mathcal{O}$ and velocity $\dot{\vec{q}}\in \mathbb{R}^\mathcal{O}$, denoted by 
$\vec{D}=\{\{t_{n,w}, \vec{q}_{n,w}, \dot{\vec{q}}_{n,w} \}_{n=1}^{N}\}_{w=1}^{W}$, where $N$ and $W$ represent the trajectory length and number of demonstrations, respectively. Similarly to many previous works, e.g., \cite{calinon2016, huang2017}, we use GMM to model the probabilistic distribution of $\vec{D}$, leading to
\begin{equation}
\mathcal{P}(t,\vec{\xi}) \sim \sum_{c=1}^{C} \pi_c \mathcal{N}(\vec{\mu}_c,\vec{\Sigma}_c),
\label{equ:gmm}
\end{equation}
where $\vec{\xi}=[\vec{q}^\trsp \,  \dot{\vec{q}}^\trsp ]^\trsp\in \mathbb{R}^{2\mathcal{O}}$, $C$ is the number of Gaussian components, $\pi_c$,
$\vec{\mu}_c=\left[\begin{matrix}
\vec{\mu}_{t,c} \\ \vec{\mu}_{\xi,c}
\end{matrix}\right]$ and
$\vec{\Sigma}_c=\left[\begin{matrix}
\vec{\Sigma}_{tt,c} & \vec{\Sigma}_{t\xi,c} \\ \vec{\Sigma}_{\xi t,c} & \vec{\Sigma}_{\xi \xi,c}
\end{matrix}\right]$ respectively denote prior probability, mean and covariance of the $c$-\emph{th} Gaussain component. Furthermore, with the parameters of GMM in (\ref{equ:gmm}), a probabilistic reference trajectory can be retrieved via Gaussian mixture regression (GMR) , giving $=\{t_{n}, \hat{\vec{\xi}}_n\}_{n=1}^{N}$ with $ \mathcal{P}(\hat{\vec{\xi}}_n|t_n)\sim \mathcal{N}(\hat{\vec{\mu}}_{n}, \hat{\vec{\Sigma}}_{n})$. Here, $\vec{D}_r=\{t_{n}, \hat{\vec{\mu}}_{n}, \hat{\vec{\Sigma}}_{n}\}_{n=1}^{N}$ can be viewed as a probabilistic representation of demonstrations $\vec{D}$.
Please refer to \cite{calinon2016,huang2017} for more details on the modeling of demonstrations.

\subsection{Problem Description of EKMP \label{subsec:ekmp:problem}}

We start with the unconstrained imitation learning method KMP to learn the reference trajectory $\vec{D}_r$.
Specifically, we parameterize $\vec{\xi}(t)$ as
\begin{equation}
\vec{\xi}(t)\!\!=\!\!\left[\!\begin{array}{c}
\vec{q}(t) \\ \dot{\vec{q}}(t)
\end{array}\!\right]
=\vec{\Theta}^{\trsp}(t)\vec{w}
\label{equ:para:traj}
\end{equation}
where the basis function matrix  $\vec{\Theta}(t)\in \mathbb{R}^{B\mathcal{O} \times 2\mathcal{O}}$ is
\begin{equation}
\vec{\Theta}(t)=[\vec{I}_{\mathcal{O}} \otimes \vec{\varphi}(t) \quad \ \vec{I}_{\mathcal{O}} \otimes \dot{\vec{\varphi}}(t)] 
\end{equation}
with $\vec{\varphi}(t) \in \mathbb{R}^{\mathcal{B}}$ being a basis function vector. `$\otimes$' denotes Kronecker product and $\vec{w}\in\mathbb{R}^{\mathcal{B}\mathcal{O}}$ denotes the trajectory parameter vector. As suggested in KMP \cite{huang2017}, the optimal $\vec{w}$ from learning demonstrations can be obtained by maximizing the posterior $\prod_{n=1}^{N}\mathcal{P}(\vec{\xi}(t_n)|\hat{\vec{\mu}}_n,\hat{\vec{\Sigma}}_n)$, which is equivalent to the  problem of minimizing
\begin{equation}
\begin{aligned}
\mathcal{U}_{IL}(\vec{w})= \sum_{n=1}^{N}\frac{1}{2}(\vec{\Theta}^{\trsp}(t_n)\vec{w}-\hat{\vec{\mu}}_n)^{\trsp} \hat{\vec{\Sigma}}_n^{-1}
(\vec{\Theta}^{\trsp}(t_n)\vec{w}-\hat{\vec{\mu}}_n) \\
+ \frac{1}{2}\lambda \vec{w}^{\trsp}\vec{w},
\end{aligned}
\label{equ:imi:cost}
\end{equation}
where $\frac{1}{2}\lambda \vec{w}^{\trsp}\vec{w}$ with $\lambda>0$ serves as the regularization term to alleviate the over-fitting issue. 

In order to cope with the obstacle avoidance problem, we employ the cost function from \cite{marinho2016functional} and define it using the parametric trajectory in (\ref{equ:para:traj}), i.e.,
\begin{equation}
\mathcal{U}_{obs}(\vec{w})=\sum_{m=1}^{M} c\bigl(\vec{x}(\vec{q} (\tilde{t}_m),u_m)\bigr),
\label{equ:obs:cost}
\end{equation}
where $M$ denotes the number of points used for calculating the obstacle avoidance cost. $\vec{x}(\vec{q} (\tilde{t}_m),u_m) \in \mathbb{R}^3$ corresponds to the workspace position of the $u_m$-\emph{th} body point\footnote{The robotic arm is assumed to be covered by a series of body points, please refer to \cite{ratliff,marinho2016functional} for the  details.} at joint position $\vec{q}(\tilde{t}_m)$. $c(\cdot)$ is the cost function. Here, we choose $u_m$ as the nearest body point to the obstacle at time $\tilde{t}_m$. Similar treatment was also implemented in \cite{park2008movement}. 
Note that other cost functions developed in \cite{osa2017guiding, ratliff, kalakrishnan2011stomp} can be used as well. 

Now, we formulate the problem of EKMP as
\begin{equation}
\begin{aligned}
\argmin_{\vec{w}}& \, \ \, \mathcal{U}(\vec{w})=\mathcal{U}_{IL}(\vec{w})+\lambda_{obs}\mathcal{U}_{obs}(\vec{w})\\
\quad\quad \textbf{s.t.} ,
&
\begin{array}{cc}
{g}_{n,1}(\vec{\Theta}^{\trsp}(t_n)\vec{w})& \geq 0 \\
{g}_{n,2}(\vec{\Theta}^{\trsp}(t_n)\vec{w})&\geq 0\\
\vdots&  \\
{g}_{n,F}(\vec{\Theta}^{\trsp}(t_n)\vec{w})& \geq 0 \\
\end{array},
\forall n \in\{1,2,\ldots,N\},
\end{aligned}
\label{equ:ini:problem}
\end{equation}
where 
$\lambda_{obs}>0$ is a constant and
$F$ denotes the number of nonlinear constraints at each time step\footnote{The number of trajectory points imposed with constraints is not necessarily the same as that of demonstrations, depending on tasks at hand.}.
$g_{n,f}$ represents the $f$-\emph{th} nonlinear constraint acting on 
$\vec{\xi}(t_n)=\vec{\Theta}^{\trsp}(t_n)\vec{w}$.
 
The objective function $\mathcal{U}(\vec{w})$ consisting of imitation learning and obstacle avoidance costs shares a similar spirit with \cite{osa2017guiding, shyam2019improving,rana2017towards}.
However, differing from \cite{osa2017guiding,rana2017towards} that  optimize discrete waypoints of a trajectory,
we propose to learn the trajectory parameter $\vec{w}$. A straightforward advantage of trajectory parameterization is that the optimal solution $\vec{w}$ of (\ref{equ:ini:problem}) can be used to generate trajectories $\vec{q}(t)$ and $\dot{\vec{q}}(t)$ simultaneously via (\ref{equ:para:traj}) which strictly satisfy the derivative relationship---a crucial requirement in many time-contact tasks. To take the striking task as an example, the racket is demanded to strike an incoming ball at a desired position with a desired velocity.  If the derivative relationship can not be respected when planning position and velocity trajectories, the racket's actual velocity (i.e., derivative of position) at the striking position will be different from the desired striking velocity, thus failing to strike the ball properly. Note that the same parameterization was also used in \cite{paraschos} for learning unconstrained motions, which was extended in \cite{shyam2019improving} to deal with the obstacle avoidance problem, whereas both \cite{shyam2019improving,paraschos} relied on explicit design of basis functions, while, as shall be seen later, in EKMP basis functions are alleviated by virtue of kernelization.
It is worthwhile to mention that
hard constraints are not considered in \cite{osa2017guiding, shyam2019improving,rana2017towards,paraschos}, which however is an essential goal in EKMP.

\subsection{EKMP \label{subsec:ekmp}}
Similarly to \cite{ratliff,marinho2016functional}, 
we approximate $\mathcal{U}(\vec{w})$ using its first-order Taylor approximation. Formally, we respectively approximate $\mathcal{U}_{IL}(\vec{w})$  in (\ref{equ:imi:cost})  and $\mathcal{U}_{obs}(\vec{w})$ in (\ref{equ:obs:cost}) using the current trajectory parameter $\vec{w}^{c}$, giving
\begin{equation}
\begin{aligned}
&\mathcal{U}_{IL}(\vec{w})=\mathcal{U}_{IL}(\vec{w}^c)+(\bigtriangledown_{\vec{w}}\mathcal{U}_{IL}(\vec{w})|_{\vec{w}=\vec{w}^c} )^{\trsp} (\vec{w}-\vec{w}^c)\\
&=\mathcal{U}_{IL}(\vec{w}^c) \!+\! \left(\vec{\Phi}\vec{\Sigma}^{-1}\vec{\Phi}^{\trsp}\vec{w}^c \!-\! \vec{\Phi}\vec{\Sigma}^{-1}\vec{\mu}+\lambda \vec{w}^c \right)^{\trsp}  \!\!(\vec{w}-\vec{w}^c)
\end{aligned}
\label{equ:imi:appro}
\end{equation}
with
\begin{equation}
\begin{aligned}
&\vec{\Phi}=[
\vec{\Theta}(t_1) \ \vec{\Theta}(t_2) \ \cdots \ \vec{\Theta}(t_N)
],\\
&\vec{\Sigma}=\mathrm{blockdiag}(\hat{\vec{\Sigma}}_1, \ \hat{\vec{\Sigma}}_2, \ \ldots, \ \hat{\vec{\Sigma}}_N), \quad \\
&{\vec{\mu}}=[
\hat{\vec{\mu}}_1^{\trsp} \ \hat{\vec{\mu}}_2^{\trsp} \ \cdots \ \hat{\vec{\mu}}_N^{\trsp}
]^{\trsp}, \\
\end{aligned}
\label{equ:notations:define:imi}
\end{equation}
and
\begin{equation}
\begin{aligned}
&\mathcal{U}_{obs}(\vec{w})=\mathcal{U}_{obs}(\vec{w}^c)+(\bigtriangledown_{\vec{w}}\mathcal{U}_{obs}(\vec{w})|_{\vec{w}=\vec{w}^c} )^{\trsp} (\vec{w}-\vec{w}^c)\\
&=\mathcal{U}_{obs}(\vec{w}^c)+ \bigl(\widetilde{\vec{\Phi}}\vec{H}^c \bigr)^{\trsp} (\vec{w}-\vec{w}^c)
\end{aligned}
\label{equ:obs:appro}
\end{equation}
with
\begin{equation}
\begin{aligned}
&\widetilde{\vec{\Phi}}=[
\vec{\Theta}(\tilde{t}_1) \ \vec{\Theta}(\tilde{t}_2) \ \cdots \ \vec{\Theta}(\tilde{t}_M)],\\
&\vec{H}^c=[{\vec{H}_1^c}^\trsp \ {\vec{H}^c_2}^\trsp \cdots  {\vec{H}^c_M}^\trsp ]^\trsp,\\
&\vec{H}^c_m=\left[\begin{array}{cc}
\vec{J}^\trsp(\tilde{t}_m,u_m) \bigtriangledown_{\vec{x}}c(\vec{x})|_{\vec{x}=\vec{x}(\vec{q} (\tilde{t}_m),u_m)}\\
\vec{0}
\end{array}\right] ,
\end{aligned}
\label{equ:notations:define:obs}
\end{equation}
where $\vec{J}(\tilde{t}_m, u_m)=\frac{\partial \vec{x}(\vec{q},u_m)}{\partial \vec{q}}|_{\vec{q}=\vec{q}(\tilde{t}_m)}$ corresponds to the Jacobian matrix of the $u_m$-\emph{th} body point when joint configuration is $\vec{q} (\tilde{t}_m)$. The zero vector ($\mathcal{O}$-dimensional) in $\vec{H}^c_m$ is due to the fact that $\mathcal{U}_{obs}$ in (\ref{equ:obs:cost}) is independent of joint velocity. In the case of only optimizing joint positions, as done in \cite{marinho2016functional}, this zero vector in $\vec{H}^c_m$ will not exist . 

Furthermore, nonlinear constraints $g_{n,f}(\vec{\Theta}^{\trsp}(t_n)\vec{w})$ can be approximated by
\begin{equation}
\begin{aligned}
	g_{n,f}(&\vec{\Theta}^{\trsp}(t_n)\vec{w})=g_{n,f}(\vec{\Theta}^{\trsp}(t_n)\vec{w}^c) \\
	 &+ (\bigtriangledown_{\vec{\xi}} g_{n,f}(\vec{\xi})|_{\vec{\xi}=\vec{\Theta}^{\trsp}(t_n)\vec{w}^c})^{\trsp} \vec{\Theta}^{\trsp}(t_n)(\vec{w}-\vec{w}^c).
\end{aligned}
\label{equ:constraint:appro}
\end{equation}
Here, the linearization is a typical way to deal with nonlinear constraints \cite{varol2012constrained}.
Let us write
\begin{equation}
\begin{aligned}
\vec{\xi}^{c}_{n}&=\vec{\Theta}^{\trsp}(t_n)\vec{w}^c, \\
\bigtriangledown g_{n,f}(\vec{\xi}^{c}_{n})&=\bigtriangledown_{\vec{\xi}} g_{n,f}(\vec{\xi})|_{\vec{\xi}=\vec{\Theta}^{\trsp}(t_n)\vec{w}^c},
\label{equ:pred:cur}
\end{aligned}
\end{equation}
where $\vec{\xi}^{c}_{n}$ denotes the predicted trajectory point at time $t_n$ with the current parameter $\vec{w}^c$. Using Lagrange multipliers $\vec{\alpha}=[\alpha_{1,1},\alpha_{1,2},\ldots,\alpha_{1,F}, \ldots , \alpha_{N,1},\alpha_{N,2},\ldots,\alpha_{N,F}]^\trsp$ with $\alpha_{n,f}\geq 0$, we can transform the problem in (\ref{equ:ini:problem}) as
\begin{equation}
\begin{aligned}
L(\vec{w},\vec{\alpha})\!=\!\biggl(\!\vec{\Phi}\vec{\Sigma}^{-1}(\vec{\Phi}^{\trsp}\vec{w}^c \!-\! \vec{\mu})\!+\! \lambda \vec{w}^c \!+\! \lambda_{obs}\widetilde{\vec{\Phi}}\vec{H}^c\! \biggr)^{\trsp} (\vec{w}\!-\!\vec{w}^c)\\ 
\!-\! \sum_{n=1}^{N}\! \sum_{f=1}^{F}\! \alpha_{n,f} \biggl(\!g_{n,f}(\vec{\xi}^{c}_{n}) \!+\! \bigl(\bigtriangledown g_{n,f}(\vec{\xi}^{c}_{n})\bigr)^{\trsp} \!\! \vec{\Theta}^{\trsp}(t_n)(\vec{w}-\vec{w}^c) \! \biggr)\\
+\frac{1}{2}\beta  (\vec{w}-\vec{w}^c)^\trsp(\vec{w}-\vec{w}^c),
\end{aligned}
\label{equ:lagrange}
\end{equation}
where $\mathcal{U}_{IL}(\vec{w}^c)$ in (\ref{equ:imi:appro}) and $\mathcal{U}_{obs}(\vec{w}^c)$ in (\ref{equ:obs:appro}) are ignored since they are constant at the current iteration. The last term $\frac{1}{2}\beta  (\vec{w}-\vec{w}^c)^\trsp(\vec{w}-\vec{w}^c)$ with $\beta>0$ is added to mitigate an overlarge update of  $\vec{w}$. The regularization
has also been studied in \cite{osa2017guiding,ratliff,mukadam2016gaussian,marinho2016functional}, but the regularization terms in \cite{osa2017guiding,ratliff,mukadam2016gaussian} were defined on trajectory waypoints while in \cite{marinho2016functional} it was defined in RHKS.

By setting the partial derivative of  $L(\vec{w},\vec{\alpha})$ w.r.t. $\vec{w}$ as zero, we have
\begin{equation}
\begin{aligned}
\vec{w}=(1-\frac{\lambda}{\beta})\vec{w}^{c}-\frac{1}{\beta}\bigl(\vec{\Phi}&\vec{\Sigma}^{-1}(\vec{\Phi}^{\trsp}\vec{w}^c \!-\! \vec{\mu}) \\
 &+\lambda_{obs}\widetilde{\vec{\Phi}}\vec{H}^c - \vec{\Phi} {\vec{G}}^{c} \vec{\alpha} \bigr)
\end{aligned}
\label{equ:w:update}
\end{equation}
with 
\begin{equation}
\begin{aligned}
&\vec{G}^c_n=[\bigtriangledown g_{n,1}(\vec{\xi}^{c}_{n}) \ \bigtriangledown g_{n,2}(\vec{\xi}^{c}_{n}) \ \ldots \ \bigtriangledown g_{n,F}(\vec{\xi}^{c}_{n})] ,\\
&\vec{G}^c=\mathrm{blockdiag}(\vec{G}^c_1,\ \vec{G}^c_2, \ \ldots, \ \vec{G}^c_N). \\
\end{aligned}
\label{equ:notations:define:constraint:gradient}
\end{equation}
Then, we substitute (\ref{equ:w:update}) into (\ref{equ:lagrange}), yielding
\begin{equation}
\begin{aligned}
&\widetilde{L}(\vec{\alpha})\!= \! - \frac{1}{2 \beta}\vec{\alpha}^{\trsp} {{\vec{G}}^{c}}^{\trsp}\vec{\Phi}^\trsp \vec{\Phi} {\vec{G}}^{c} \vec{\alpha}
-{\vec{Q}^{c}}^{\trsp} \vec{\alpha}  \\
&\!\!+ \!\!\frac{1}{\beta}\bigl((\vec{\Phi}^\trsp\vec{w}^{c}\!\!-\!\!\vec{\mu})^\trsp \vec{\Sigma}^{-1}\vec{\Phi}^\trsp \!\!+\!\! \lambda {\vec{w}^c}^{\trsp} \!\!+\! \!\lambda_{obs}{\vec{H}^c}^\trsp \widetilde{\vec{\Phi}}^\trsp \bigr) \vec{\Phi} {\vec{G}}^{c} \vec{\alpha}
\label{equ:alpha:function}
\end{aligned}
\end{equation}
with 
\begin{equation}
\begin{aligned}
&\vec{Q}^{c}_n=[g_{n,1}(\vec{\xi}^{c}_{n}) \ g_{n,2}(\vec{\xi}^{c}_{n}) \ \ldots \ g_{n,F}(\vec{\xi}^{c}_{n})]^{\trsp}, \\
&{\vec{Q}}^c=[\vec{Q}_1^\trsp \ \vec{Q}_2^\trsp \ldots \vec{Q}_N^\trsp]^{\trsp}.
\end{aligned}
\label{equ:notations:define:constraint:value}
\end{equation}
Furthermore, we can kernelize (\ref{equ:alpha:function}) by using the kernel trick $\vec{\varphi}(t_i)^{\trsp} \vec{\varphi}(t_j)=k(t_i,t_j)$, where $k(\cdot,\cdot)$ is a kernel function. The kernelized form of (\ref{equ:alpha:function}) is
\begin{equation}
\begin{aligned}
&\widetilde{L}(\vec{\alpha})\!= \! - \frac{1}{2 \beta}\vec{\alpha}^{\trsp} {{\vec{G}}^{c}}^{\trsp}\vec{K} {\vec{G}}^{c} \vec{\alpha}
-{\vec{Q}^{c}}^{\trsp} \vec{\alpha}  \\
&+ \! \frac{1}{\beta}\bigl(({\vec{\xi}}^c\!-\!\vec{\mu})^\trsp \vec{\Sigma}^{-1}\vec{K} \!+\! \lambda {{\vec{\xi}}^c}^{\trsp} \!\!+\! \lambda_{obs}{\vec{H}^c}^\trsp \widetilde{\vec{K}} \bigr) {\vec{G}}^{c} \vec{\alpha},
\label{equ:alpha:function:kernel}
\end{aligned}
\end{equation}
where ${\vec{\xi}}^c$ represents the predicted trajectory, i.e.,
\begin{equation}
{\vec{\xi}}^c=\vec{\Phi}^\trsp\vec{w}^{c}=[{\vec{\xi}^c_1}^\trsp {\vec{\xi}^c_2}^\trsp \ldots {\vec{\xi}^c_N}^\trsp ]^\trsp.
\label{equ:pred:cur:entire:traj}
\end{equation}
$\vec{K}$ denotes a $N\times N$ block matrix defined by
\begin{equation}
\vec{K}
\!\!=\!\!\vec{\Phi}^{\trsp}\vec{\Phi}\!\!=\!\!\left[\begin{matrix} 
\vec{k}(t_1, t_1) & \vec{k}(t_1, t_2) & \cdots &\vec{k}(t_1, t_N) \\
\vec{k}(t_2, t_1) & \vec{k}(t_2, t_2) & \cdots &\vec{k}(t_2, t_N) \\
\vdots & \vdots &  \ddots & \vdots \\
\vec{k}(t_N, t_1) & \vec{k}(t_N, t_2) & \cdots &\vec{k}(t_N, t_N) \\
\end{matrix}\right]
\label{equ:K:matrix}
\end{equation}
with
\begin{equation}
\vec{k}(t_i,t_j)\!=\! \vec{\Theta}({t_i})^{\trsp}\vec{\Theta}({t_j})\!\!=\!\!
\left[ \begin{matrix} k_{tt}(i,j)\vec{I}_{\mathcal{O}} \!&\! k_{td}(i,j)\vec{I}_{\mathcal{O}}\\
k_{dt}(i,j)\vec{I}_{\mathcal{O}} \!&\! k_{dd}(i,j)\vec{I}_{\mathcal{O}} \\
\end{matrix} \right],
\label{equ:kernel:matrix:time}
\end{equation}
where
$
k_{tt}(i,j)=k(t_i,t_j),
k_{td}(i,j)=\frac{k(t_i,t_j+\delta)-k(t_i,t_j)}{\delta},
k_{dt}(i,j)=\frac{k(t_i+\delta,t_j)-k(t_i,t_j)}{\delta}, 
k_{dd}(i,j)=\frac{k(t_i\!+\delta, t_j+\delta) -k(t_i+\delta, t_j) -k(t_i,t_j+\!\delta) +k(t_i,t_j)}{{\delta}^{2}}. 
$
Here, $\delta>0$ denotes a small constant. Please refer to KMP \cite{huang2017} for more details of the kernelization. Similarly, $\widetilde{\vec{K}}=\vec{\Phi}^{\trsp}\widetilde{\vec{\Phi}}$ corresponding to  a $N\times M$ block matrix can be obtained.

Since in (\ref{equ:alpha:function:kernel}) the quadratic coefficient $- \frac{1}{2 \beta} {{\vec{G}}^{c}}^{\trsp}\vec{K}{\vec{G}}^{c}$ is a symmetric and negative definite matrix, we can resort to  \emph{quadratic programming} (QP) to find the optimal $\vec{\alpha}^{c}$ at the current iteration maximizing (\ref{equ:alpha:function:kernel}) while satisfying the constraint $\vec{\alpha}\geq\vec{0}$. 
Therefore, for an arbitrary time $t$, its corresponding trajectory point can be determined via  (\ref{equ:w:update}) with $\vec{\alpha}=\vec{\alpha}^{c}$, i.e.,
\begin{equation}
\begin{aligned}
&\vec{\xi}(t)=\left[\begin{array}{c}
\vec{q}(t) \\ \dot{\vec{q}}(t)
\end{array}\right]
=\vec{\Theta}^{\trsp}(t)\vec{w} \\
&=(1-\frac{\lambda}{\beta})\vec{\Theta}^{\trsp}(t)\vec{w}^{c}-\frac{1}{\beta}\bigl(\vec{\Theta}^{\trsp}(t)\vec{\Phi}\vec{\Sigma}^{-1}(\vec{\Phi}^{\trsp}\vec{w}^c \!-\! \vec{\mu}) \\ 
&+\lambda_{obs}\vec{\Theta}^{\trsp}(t)\widetilde{\vec{\Phi}}{\vec{H}^c} - \vec{\Theta}^{\trsp}(t)\vec{\Phi} {\vec{G}}^{c} \vec{\alpha}^c \bigr).
\end{aligned}
\label{equ:prediction:update}
\end{equation}
Again, by employing the kernel trick, we have
\begin{equation}
\begin{aligned}
\vec{k}(t)&=\vec{\Theta}^{\trsp}(t)\vec{\Phi}
=\left[ 
\vec{k}(t, t_1) \ \vec{k}(t, t_2) \ \cdots \ \vec{k}(t, t_N)\right], \\
\widetilde{\vec{k}}(t)&=\vec{\Theta}^{\trsp}(t)\widetilde{\vec{\Phi}}=\left[ 
\vec{k}(t, \tilde{t}_1) \ \vec{k}(t, \tilde{t}_2) \ \cdots \ \vec{k}(t, \tilde{t}_M)\right].
\end{aligned}
\label{equ:kernel:matrix:pre}
\end{equation}
Let us write $\vec{\xi}^{c}(t)=\vec{\Theta}^{\trsp}(t)\vec{w}^c$, the updated \emph{trajectory function} corresponding to (\ref{equ:prediction:update}) becomes
\begin{equation}
\begin{aligned}
\vec{\xi}(t)=(1-\frac{\lambda}{\beta})\vec{\xi}^{c}(t) - \frac{1}{\beta}\biggl(\vec{k}(t)\vec{\Sigma}^{-1}({\vec{\xi}}^c -\vec{\mu})  \\
+\lambda_{obs}\widetilde{\vec{k}}(t)\vec{H}^c -\vec{k}(t) {\vec{G}}^{c} \vec{\alpha}^c \biggr).
\end{aligned}
\label{equ:update:rule:kernel}
\end{equation}
Thus, we have obtained a kernelized rule for updating the trajectory function. 

\subsection{Implementation of EKMP \label{subsec:ekmp:imple}}
To illustrate  how the trajectory function can be updated using (\ref{equ:update:rule:kernel}), we here explain the implementation details of the first two iterations. 
In practice, we can use vanilla KMP to generate an initial trajectory before applying the iteration rule in (\ref{equ:update:rule:kernel}), since KMP provides an analytical solution for imitation learning. Formally, the initial trajectory function from vanilla KMP is
\begin{equation}
\vec{\xi}^{(0)}(t)=\vec{k}(t)\underbrace{( \vec{K} +\lambda \vec{\Sigma} )^{-1} \vec{\mu}}_{\vec{\gamma}^({0})},
\label{equ:kmp}
\end{equation}
which can be used to predict  trajectory points at time steps $\{t_n\}_{n=1}^N$, i.e., $\vec{\xi}^{(0)}=\vec{K}\vec{\gamma}^{(0)}$.
Given $\vec{\xi}^{(0)}$, we can calculate $\vec{H}^{(0)}$, $\vec{G}^{(0)}$ and $\vec{Q}^{(0)}$ via (\ref{equ:notations:define:obs}), (\ref{equ:notations:define:constraint:gradient}) and (\ref{equ:notations:define:constraint:value}), respectively.  Furthermore, the optimal $\vec{\alpha}^{(0)}$ 
can be determined by applying the QP optimization to (\ref{equ:alpha:function:kernel}). Denote $\vec{e}^{(0)}=\vec{\Sigma}^{-1}({\vec{\xi}}^{(0)} -\vec{\mu}) 
-{\vec{G}}^{(0)} \vec{\alpha}^{(0)}$,  the updated trajectory function after the first iteration becomes
\begin{equation}
\begin{aligned}
\vec{\xi}^{(1)}(t)=\vec{k}(t)\underbrace{ \bigl( (1-\frac{\lambda}{\beta})\vec{\gamma}^{(0)} - \frac{1}{\beta}\vec{e}^{(0)}\bigr)}_{\vec{\gamma}^{(1)}} 
-\widetilde{\vec{k}}(t)\underbrace{(\frac{\lambda_{obs}}{\beta}\vec{H}^{(0)} )}_{\vec{\rho}^{(1)}},
\end{aligned}
\label{equ:update:rule:kernel:exmaple:first:iter}
\end{equation}
where the kernel matrices $\vec{K}$, $\tilde{\vec{K}}$, $\vec{k}$ and $\tilde{\vec{k}}$ are determined using (\ref{equ:K:matrix}), (\ref{equ:kernel:matrix:time}) and (\ref{equ:kernel:matrix:pre}), $\vec{\Sigma}$ and $\vec{\mu}$ are calculated via (\ref{equ:notations:define:imi}). 

In a similar manner, we can predict $\vec{\xi}^{(1)}=\vec{K}\vec{\gamma}^{(1)}-\widetilde{\vec{K}}\vec{\rho}^{(1)}$, $\vec{H}^{(1)}$, $\vec{G}^{(1)}$, $\vec{Q}^{(1)}$, $\vec{\alpha}^{(1)}$ and $\vec{e}^{(1)}=\vec{\Sigma}^{-1}({\vec{\xi}}^{(1)}-\vec{\mu}) 
-{\vec{G}}^{(1)} \vec{\alpha}^{(1)}$, respectively. Thus,  the updated trajectory function after the second iteration is
\begin{equation}
\begin{aligned}
\vec{\xi}^{(2)}(t)=\vec{k}(t)&\underbrace{ \bigl( (1-\frac{\lambda}{\beta})\vec{\gamma}^{(1)} - \frac{1}{\beta} \vec{e}^{(1)}\bigr)}_{\vec{\gamma^{(2)}}} \\
&-\widetilde{\vec{k}}(t)\underbrace{\bigl((1-\frac{\lambda}{\beta})\vec{\rho}^{(1)}+\frac{\lambda_{obs}}{\beta}\vec{H}^{(1)} \bigr)}_{\vec{\rho}^{(2)}}.
\end{aligned}
\label{equ:update:rule:kernel:exmaple:second:iter}
\end{equation}
Observing (\ref{equ:update:rule:kernel:exmaple:first:iter})--(\ref{equ:update:rule:kernel:exmaple:second:iter}), we can see that only $\vec{\gamma}^{(n)}$ and $\vec{\rho}^{(n)}$ need to be updated iteratively till convergence. Let us denote the number of total iterations as $n^*$, the \emph{optimal trajectory function} for the problem in (\ref{equ:ini:problem}) will be
\begin{equation}
\vec{\xi}^{*}(\cdot)=\vec{k}(\cdot)\vec{\gamma}^{(n^*)}-\widetilde{\vec{k}}(\cdot)\vec{\rho}^{(n^*)},
 \end{equation}
which is capable of predicting the corresponding trajectory point (including joint position and velocity) at arbitrary time. 
The entire EKMP algorithm is summarized in Algorithm~\ref{alg:ekmp}.

\begin{algorithm}
	\textbf{Initialization}\\
	Set $\lambda$, $\beta$, $\lambda_{obs}$, $k(\cdot,\cdot)$ and define $\{\{\vec{g}_{n,f}\}_{n=1}^{N}\}_{f=1}^{F}$	 \\
	Extract reference trajectory $\vec{D}_r$ from demonstrations\\
	Calculate $\vec{\mu}$ and $\vec{\Sigma}$ via (\ref{equ:notations:define:imi}), $\vec{K}$ via (\ref{equ:K:matrix}) and $\widetilde{\vec{K}}$  \\
	Calculate $\!\vec{\gamma}^{(0)}$ via (\ref{equ:kmp}) and $\vec{\xi}^{(0)}=\vec{K}\vec{\gamma}^{(0)}$ \\
	Calculate $\vec{H}^{(0)},\vec{G}^{(0)},\vec{Q}^{(0)}$ via  (\ref{equ:notations:define:obs}), (\ref{equ:notations:define:constraint:gradient}) and  (\ref{equ:notations:define:constraint:value}) \\
	Optimize $\vec{\alpha}^{(0)}$ in (\ref{equ:alpha:function:kernel}) using QP  and calculate $\vec{e}^{(0)}$ \\
	Set a maximal iteration number $N^{*}$, tolerance error $\epsilon$\\
	Set $\vec{\rho}^{(0)}=\vec{0}$, $iter=0$ \\
	\While{$iter<N^{*}$}{
		$\vec{\gamma}^{(iter+1)} \leftarrow  (1-\frac{\lambda}{\beta})\vec{\gamma}^{(iter)} - \frac{1}{\beta}\vec{e}^{(iter)}$ \\
		$\vec{\rho}^{(iter+1)} \leftarrow (1-\frac{\lambda}{\beta})\vec{\rho}^{(iter)}+\frac{\lambda_{obs}}{\beta}\vec{H}^{(iter)} $\\
		Calculate $\vec{\xi}^{(iter+1)}=\vec{K}\vec{\gamma}^{(iter+1)}-\widetilde{\vec{K}}\vec{\rho}^{(iter+1)}$\\
		Calculate $\vec{H}^{(iter+1)}$, $\vec{G}^{(iter+1)}$, $\vec{Q}^{(iter+1)}$ \\
		Optimize  $\vec{\alpha}^{(iter+1)}$ and calculate $\vec{e}^{(iter+1)}$ \\
		${\delta}_{\gamma} \leftarrow ||\vec{\gamma}^{(iter+1)} -\vec{\gamma}^{(iter)}|| $\\
		${\delta}_{\rho} \leftarrow ||\vec{\rho}^{(iter+1)} -\vec{\rho}^{(iter)}||$\\
		\If{$||{\delta}_{\gamma} ||<\epsilon$ and $||{\delta}_{\rho} ||<\epsilon$}{
			\textbf{break}
		}
		$iter \leftarrow iter+1$ \\
	}
	\textbf{Output}: $\vec{\xi}^{*}(\cdot)=\vec{k}(\cdot)\vec{\gamma}^{(iter)}-\widetilde{\vec{k}}(\cdot)\vec{\rho}^{(iter)}$ 
	\caption{EKMP}
	\label{alg:ekmp}
\end{algorithm}

\subsection{Special Cases of EKMP \label{subsec:ekmp:extend}}
We have derived EKMP under nonlinear hard constraints. 
Now, we consider the applications of EKMP to deal with some special cases.

\textbf{Nonlinear equality constraints}: we can use inequalities to guarantee equality constraints with a tiny approximation error $\epsilon>0$, i.e.,  ${g}_{n,f}(\vec{\xi}(t_n))=0$ is enforced by ${g}_{n,f}(\vec{\xi}(t_n))+\epsilon \geq0$ and $-{g}_{n,f}(\vec{\xi}(t_n))+\epsilon \geq0$.

\textbf{Adaptations towards desired points}: 
an important feature in imitation learning is the adaptation of learned trajectories towards arbitrary desired points in terms of joint position and velocity,
including start-, via- and end-points. Assuming that
$L$ desired points, denoted by $\{\bar{t}_l,\bar{\vec{\xi}}_l\}_{l=1}^{L}$ with $\bar{\vec{\xi}}_l=[\bar{\vec{q}}_l^\trsp \ \bar{\dot{\vec{q}}}_l^\trsp]^\trsp \in \mathcal{R}^{2\mathcal{O}}$, are required in a task.
For each desired point $\{\bar{t}_l,\bar{\vec{\xi}}_l\}$ we can separately define constraints as
\begin{equation}
\begin{aligned}
{g}_{l,f}(\vec{\xi}(t_l))&=\vec{1}_f^\trsp (\vec{\xi}(t_l)-\bar{\vec{\xi}}_l)+\epsilon \geq0, \\
{g}_{l,2\mathcal{O}+f}(\vec{\xi}(t_l))&=-\vec{1}_{f}^\trsp (\vec{\xi}(t_l)-\bar{\vec{\xi}}_l)+\epsilon \geq0
\end{aligned}
\label{equ:des:point}
\end{equation}
for $\forall f\in\{1,2,\ldots,2\mathcal{O}\}$, where $\epsilon>0$ is used to control the adaptation precision, i.e., how precisely the adapted trajectory can go through the desired point $\bar{\vec{\xi}}_l$ at time $\bar{t}_l$. $\vec{1}_f\in\mathcal{R}^{2\mathcal{O}}$ is an indicative vector with all elements being zero except for its $f$-\emph{th} element.

\textbf{Linear constraints}: we can simply write
\begin{equation}
{g}_{n,f}(\vec{\xi}(t_n))=\vec{\theta}^{\trsp}\vec{\xi}(t_n)+b \geq 0,
\end{equation}
where $\vec{\theta}$ and $b$ are constant, representing coefficients of linear constraints. Note that a \textbf{plane constraint} can be ensured by $\vec{\theta}^{\trsp}\vec{\xi}(t_n)+b+\epsilon\geq0$ and $-\vec{\theta}^{\trsp}\vec{\xi}(t_n)-b+\epsilon\geq0$. In the case of \textbf{joint position limits}, for each joint $i$ with motion range $[q_i^{min}, q_i^{max}]$, its position limit can be formulated as  $\vec{1}_i^{\trsp}\vec{\xi}(t_n)-q_i^{min}\geq0$ and $-\vec{1}_i^{\trsp}\vec{\xi}(t_n)+q_i^{max}\geq0$. Similarly, \textbf{joint velocity limits} can also be imposed.

\section{Connections with State-of-the-Art Motion Planning Approaches \label{sec:connection}}
We now discuss the connections between EKMP and CHOMP \cite{ratliff} (Section~\ref{subsec:chomp:comp}),  GPMP \cite{mukadam2016gaussian} (Section~\ref{subsec:gpmp:comp}) and RKHS motion planning \cite{marinho2016functional} (Section~\ref{subsec:rkhs:comp}).

\subsection{Comparison with CHOMP \label{subsec:chomp:comp}}
Since in practice the implementation of CHOMP updates discrete waypoints of a trajectory \cite{marinho2016functional}, we discretize 
the update rule of EKMP (i.e., (\ref{equ:update:rule:kernel})) to establish the connection.
Supposing that we describe a trajectory by using its waypoints at a series of time steps $\{t_n\}_{n=1}^{N}$. By using (\ref{equ:update:rule:kernel}), we update each waypoint as
\begin{equation}
\begin{aligned}
&\underbrace{\left[\begin{array}{c}
\vec{\xi}(t_1)\\ \vec{\xi}(t_2)\\\vdots\\\vec{\xi}(t_N)\\
\end{array}\right]}_{\vec{\xi}}
=(1-\frac{\lambda}{\beta})
\underbrace{\left[\begin{array}{c}
\vec{\xi}^c(t_1)\\ \vec{\xi}^c(t_2)\\\vdots\\\vec{\xi}^c(t_N)\\
\end{array}\right] }_{\vec{\xi}^c}\\
& - \frac{1}{\beta}\bigl(\vec{K}\vec{\Sigma}^{-1}\bigl(
\vec{\xi}^c -\vec{\mu}\bigr) 
+\lambda_{obs}\widetilde{\vec{K}}\vec{H}^c -\vec{K} {\vec{G}}^{c} \vec{\alpha}^c \bigr),
\end{aligned}
\label{equ:update:rule:kernel:discrete}
\end{equation}
which can be further simplified by letting $\widetilde{\vec{K}}=\vec{K}$, yielding
\begin{equation}
\vec{\xi}=(1-\frac{\lambda}{\beta}) \vec{\xi}^c - \frac{1}{\beta}\vec{K}\bigl(\vec{\Sigma}^{-1}(\vec{\xi}^c-\vec{\mu})+\lambda_{obs} \vec{H}^c-\vec{G}^c \vec{\alpha}^c\bigr).
\label{equ:update:rule:kernel:discrete:simplify}
\end{equation}
If we let $\lambda=0$, $\vec{\mu}=\vec{0}$ and
$\vec{\alpha}^c=\vec{0}$, (\ref{equ:update:rule:kernel:discrete:simplify}) will become
\begin{equation}
\vec{\xi}=\vec{\xi}^c -  \frac{1}{\beta}\underbrace{\vec{K}\bigl(\vec{\Sigma}^{-1}\vec{\xi}^c+\lambda_{obs} \vec{H}^c\bigr)}_{update \,\, term},
\end{equation}
which shares a similar form with the update rule in CHOMP\footnote{CHOMP also considered an additional vector dealing with boundary conditions in its update rule.}. 
However, in CHOMP the coefficient of  $\vec{\xi}^c$ in the update term is $\vec{K}\vec{K}^{-1}=\vec{I}$ rather than $\vec{K}\vec{\Sigma}^{-1}$. Moreover,
$\vec{K}$ in CHOMP is defined using the finite difference matrix in order to ensure the smoothness of a trajectory, while $\vec{K}$ (i.e., (\ref{equ:K:matrix})) in EKMP is a kernel matrix. Note that hard equality constraints were addressed in \cite{ratliff}, but without considering nonlinear inequality constraints.

\subsection{Comparison with GPMP \label{subsec:gpmp:comp}}
Let us continue with (\ref{equ:update:rule:kernel:discrete:simplify}) and if we let $\lambda=0$ and neglect nonlinear hard constraints (i.e., $\vec{\alpha}^c=\vec{0}$), (\ref{equ:update:rule:kernel:discrete:simplify}) can be rewritten as
\begin{equation}
\vec{\xi}=\vec{\xi}^c - \frac{1}{\beta} \underbrace{\vec{K}\bigl(\vec{\Sigma}^{-1}(\vec{\xi}^c-\vec{\mu})+\lambda_{obs} \vec{H}^c\bigr)}_{update \,\, term},
\end{equation} 
which resembles the update rule in GPMP\footnote{A projection matrix imposed on the gradient of the obstacle avoidance cost was also considered in GPMP.}. However, in GPMP the coefficient of  $\vec{\xi}^c-\vec{\mu}$ in the update term is $\vec{K}\vec{K}^{-1}=\vec{I}$ rather than $\vec{K}\vec{\Sigma}^{-1}$. Specifically, $\vec{K}$ in GPMP is obtained from a stochastic differential equation and has a specific form related to state transition matrix as well as noise distribution. In contrast, $\vec{K}$ in EKMP is a generic kernel function (e.g., Gaussian kernel and periodic kernel), which can be chosen depending on task requirements. 

The advantage of using $\vec{\Sigma}^{-1}$ to weigh 
$\vec{\xi}^c-\vec{\mu}$ in EKMP is that the importance of demonstrations can be incorporated into the process of updating trajectories, i.e., datapoints with small covariance (large consistency implies high importance) will have larger impact on trajectory update than those with large covariance. Similar insights have been  discussed in previous  works but in the context of unconstrained imitation learning, e.g., \cite{calinon2016,huang2017}.
Note that GPMP was extended in \cite{rana2017towards} to encapsulate the prior distribution of demonstrations, where hard constraints were also neglected.

In addition, in terms of the trajectory update form, \cite{mukadam2016gaussian,rana2017towards} optimizes trajectory waypoints while in EKMP, as shown in (\ref{equ:update:rule:kernel}), the trajectory function is updated. Specifically, unlike \cite{mukadam2016gaussian,rana2017towards}, 
in EKMP the derivative relationship between the predicted joint position and velocity can be ensured over the course of optimizing (\ref{equ:ini:problem}), since the update rule (\ref{equ:update:rule:kernel}) is derived from the parametric form (\ref{equ:para:traj}) and the corresponding derivative relationship is encoded in $\vec{k}(\cdot,\cdot)$ (i.e., (\ref{equ:kernel:matrix:time})).

\begin{figure}[bt] \centering 
	\includegraphics[width=0.49\textwidth]{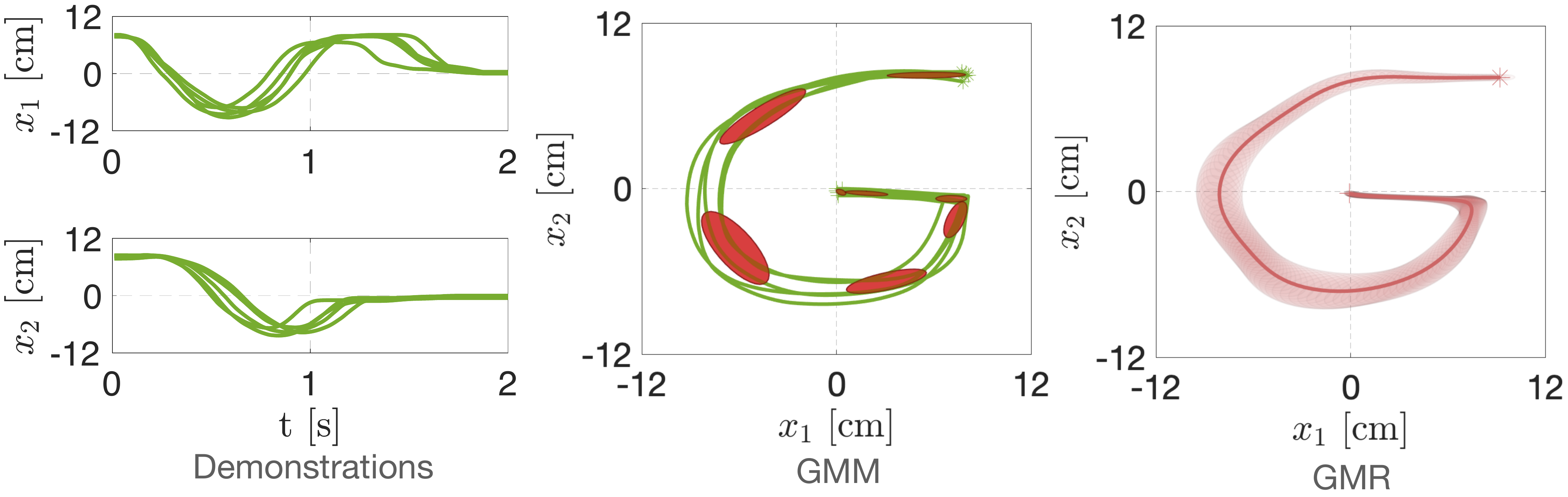}
	\vspace{-0.7cm}
	\caption{The modeling of demonstrations (green curves) using GMM and GMR, where red ellipses denote the Gaussian components in GMM, the red curve and the shaded area respectively correspond to the mean and covariance of demonstrations.}
	\label{fig:letter:demos} 
\end{figure}

\subsection{Comparison with RKHS Motion Planning \label{subsec:rkhs:comp}}
Since RKHS motion planning \cite{marinho2016functional} directly updates trajectory function, we compare it with the update rule in (\ref{equ:update:rule:kernel}). In fact, if we do not consider imitation learning (i.e., setting $\vec{\Sigma}^{-1}(\vec{\xi}^c -\vec{\mu})=\vec{0}$)  and replace nonlinear constraints with joint position and velocity limits (i.e., letting $F=2\mathcal{O}$  and $\vec{G}=-\vec{I}$), (\ref{equ:update:rule:kernel}) will become
\begin{equation}
\begin{aligned}
\vec{\xi}(\cdot)=(1-\frac{\lambda}{\beta})\vec{\xi}^{c}(\cdot) - \frac{1}{\beta}\bigl(
\lambda_{obs}\widetilde{\vec{k}}(\cdot)\vec{H}^c +\vec{k}(\cdot) \vec{\alpha}^c \bigr),
\end{aligned}
\label{equ:update:rule:kernel:simplify}
\end{equation}
which has the same form as the update rule in \cite{marinho2016functional}, except that $\vec{k}$ and $\widetilde{\vec{k}}$ have different definitions due to the simultaneous learning of $\vec{q}$ and $\dot{\vec{q}}$ in EKMP, and $\vec{H}^c$ includes a zero vector due to the optimization of $\dot{\vec{q}}$. Note that the constraints of desired starting and ending joint positions in  \cite{marinho2016functional} can be handled by inequality constraints (see Section~\ref{subsec:ekmp:extend}).

\myparagraphnp{Summary}: EKMP can be viewed as a generalization of \cite{ratliff,mukadam2016gaussian,marinho2016functional} towards imitation learning as well as nonlinear hard (in particular inequality) constraints. Moreover, EKMP can learn both joint position and velocity simultaneously while maintaining the corresponding derivative relationship even under nonlinear hard constraints and obstacle avoidance requirement. Specifically, in contrast to \cite{ratliff,mukadam2016gaussian} that optimize trajectory waypoints, EKMP updates the trajectory function, sharing the same advantages as \cite{marinho2016functional}, including fast convergence and smooth trajectories.
For the discussion of advantages of optimizing trajectory function over trajectory waypoints, please refer to \cite{marinho2016functional}.

\begin{figure*}[bt] \centering 
	\includegraphics[width=0.99\textwidth]{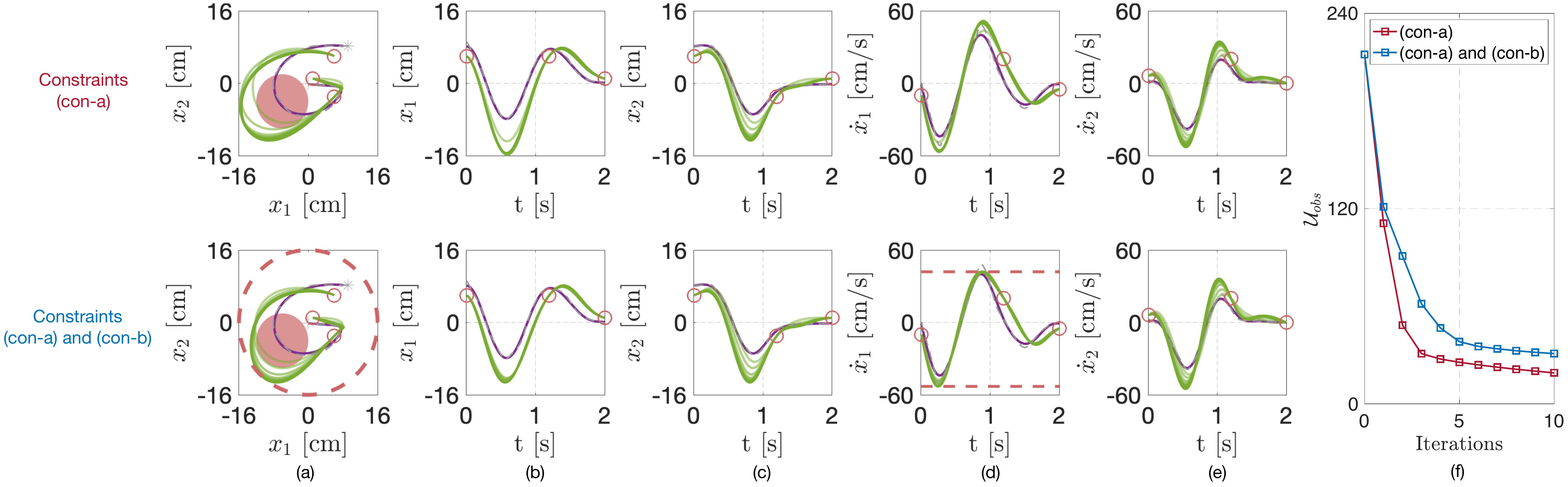}
	\vspace{-0.2cm}
	\caption{Evaluations of EKMP under different constraints. (\emph{a})--(\emph{e}): Dashed gray curves are retrieved by GMR, where `$\ast$'  and `+' denote  the start-point and end-point, respectively. Purple curves represent trajectories (obtained by vanilla KMP) used to  initialize EKMP. Evolution from light green to dark green corresponds to the updated direction over 10 iterations. Small red circles plot desired points, corresponding to the constraints \textbf{(con-a)}. The red solid circle in (\emph{a}) depicts the obstacle.  
		The dashed red circle in (\emph{a}) and dashed lines in (\emph{d}) correspond to the constraints \textbf{(con-b)}.
		(\emph{f}): Obstacle avoidance costs
	 under different constraints.}
	\label{fig:letter:evolve} 
\end{figure*}

\begin{figure}[bt] \centering 
	\includegraphics[width=0.47\textwidth]{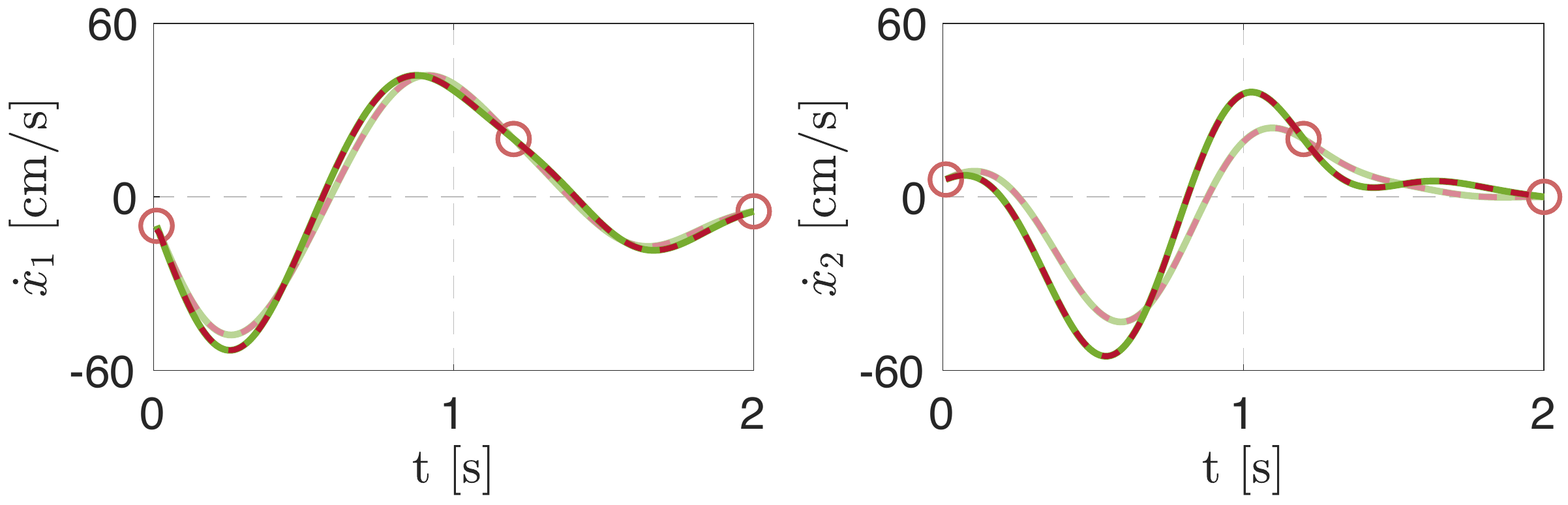}
	\vspace{-0.2cm}
	\caption{Derivative of the predicted positions (dashed red curves) and the predicted velocities (green curves) via EKMP after one iteration (light green and red) and ten iterations (dark green and red) when considering both the constraints \textbf{(con-a)} and \textbf{(con-b)} simultaneously.}
	\label{fig:letter:pos:derivative} 
\end{figure}

\begin{figure*}[bt] \centering 
	\includegraphics[width=0.94\textwidth]{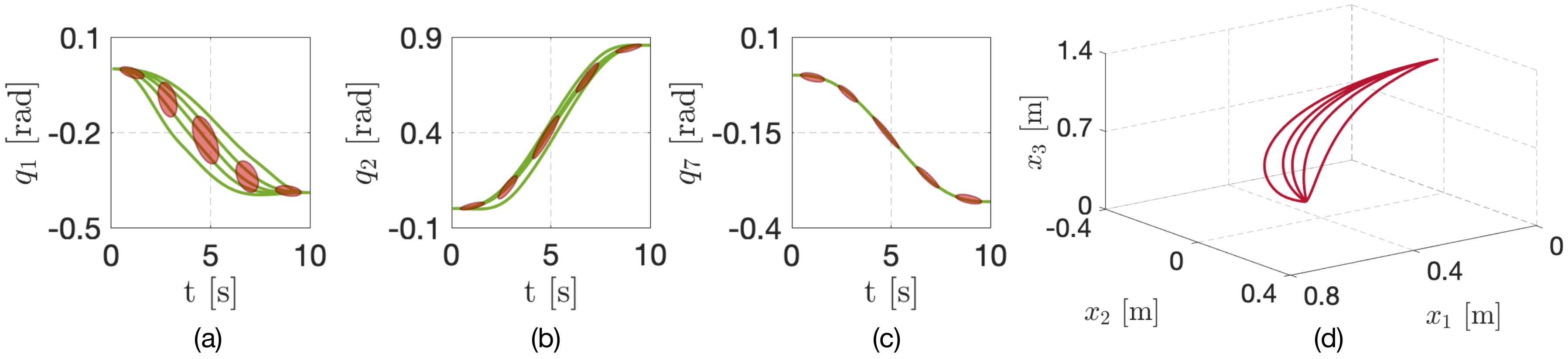}
	\caption{Demonstrations of  the reaching task. (\emph{a})--(\emph{c}) correspond to the first, second and seventh joints, respectively. (\emph{d}) plots the corresponding Cartesian trajectories of the robot's end-effector. }
	\label{fig:joint:demos} 
\end{figure*}

\begin{figure*}[bt] \centering 
	\includegraphics[width=0.98\textwidth]{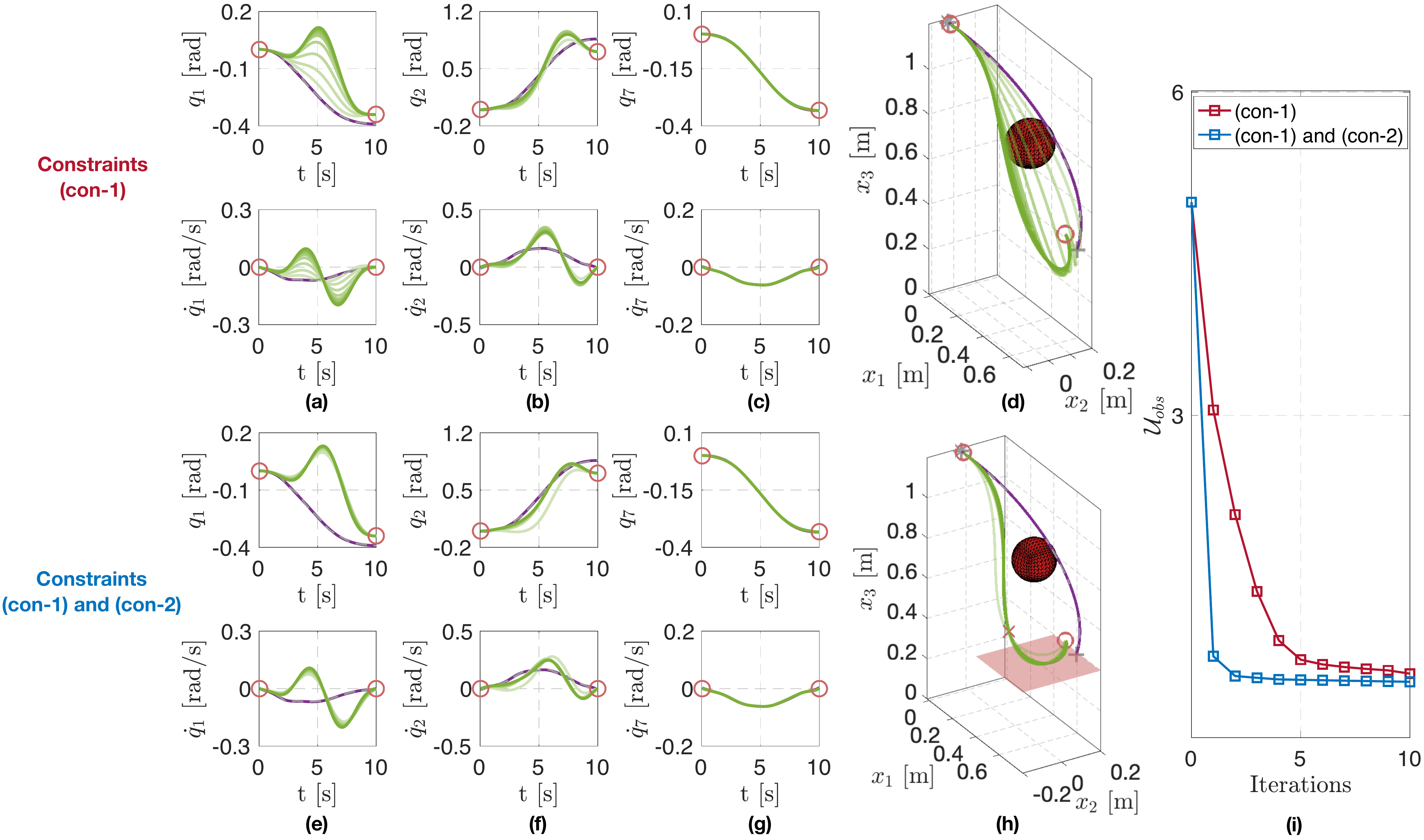}
	\caption{The planning of joint trajectories via EKMP under different constraints. 
		In (\emph{a})--(\emph{c}) and (\emph{e})--(\emph{g}), dashed gray curves are retrieved by GMR; Purple curves depict trajectories generated by vanilla KMP, which are used to initialize EKMP;  Evolution from light green to dark green shows the updated trajectories  over 10 iterations, where the color changes from light to dark as the iteration number increases;
		Red circles represent desired point constraints \textbf{(con-1)}. 
		In (\emph{d}) and (\emph{h}), gray, purple and green curves plot the corresponding end-effector trajectories of the planned joint trajectories;  Red circles depict the end-effector positions corresponding to the desired joint positions; `$\ast$'  and `+' denote  the start-point and end-point of the GMR trajectory, respectively; The red solid ball depicts the obstacle.
		In (h), `$\times$' depicts the desired Cartesian position and the red plane plots the z-direction limit, both corresponding to the constraints \textbf{(con-2)} in (\ref{equ:joint:con}).
		(\emph{f}): Obstacle avoidance costs under different constraints.}
	\label{fig:joint:eva} 
\end{figure*}

\begin{figure*}[bt] \centering 
	\includegraphics[width=0.94\textwidth]{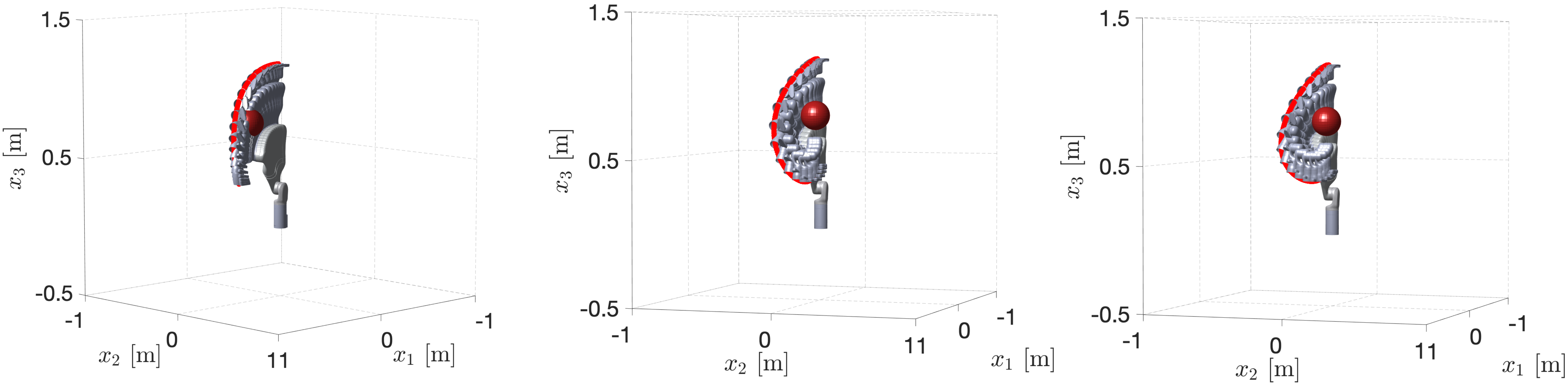}
	\caption{Different planned trajectories executed on a 7-DoF robotic arm. \textit{Left} graph shows the execution of the joint trajectory generated by vanilla KMP.  \textit{Middle} graph corresponds to the optimized joint trajectory after 10 iterations under constraints \textbf{(con-1)} while \textit{right} graph is under both \textbf{(con-1)} and \textbf{(con-2)}. }
	\label{fig:joint:robot:eva} 
\end{figure*}

\section{Evaluations \label{sec:eva}}

We evaluate EKMP through several examples, including 2-D writing tasks (Section~\ref{subsec:write}), as well as reaching tasks in a 7-DoF robotic arm (Section~\ref{subsec:robot:arm}).
The Gaussian kernel $k(t_i,t_j)=\exp(-k_h (t_i-t_j)^2)$ is used in all examples. 

\subsection{Writing Task \label{subsec:write}}
We here consider the task of writing a letter `G', where trajectories  are planned in 2-D space.
Given five demonstrations consisting of time and 2-D position, as shown in Fig.~\ref{fig:letter:demos} (\emph{left} graph), we use GMM and GMR to model their distribution and subsequently extract a probabilistic reference trajectory, see Fig.~\ref{fig:letter:demos} (\emph{middle} and \emph{right} graphs). 
Two groups of constraints are studied in this task: \\
\textbf{(con-a)}: three desired points 
in terms of 2-D position and 2-D velocity, depicted by the small red circles in Fig.~\ref{fig:letter:evolve};\\
\textbf{(con-b)}: a circle constraint and two velocity constraints
\begin{equation}
 \begin{aligned}
x_1^2+x_2^2\leq16^2,\\
-53 \leq \dot{x}_1 \leq 42.
\end{aligned}
\label{equ:letter:constraint}
\end{equation}
We first evaluate EKMP under the constraints \textbf{(con-a)} and later 
under both \textbf{(con-a)} and \textbf{(con-b)}. In both evaluations, an obstacle with a radius $r=6$\emph{cm} located at $[-6.0 \ -4.0]^\trsp$\emph{cm} should be avoided. We use the cost function $c(\cdot)$ from \cite{ratliff} for evaluating body points, i.e.,
\begin{equation}
c(d)=\left\{\begin{aligned}
r-&d+\frac{1}{2}\epsilon_d,  &d\leq r,\\
\frac{1}{2\epsilon_d}(d&-r-\epsilon_d)^2, &0<d-r\leq\epsilon_d,\\
&0, &otherwise,
\end{aligned} \right.
\end{equation}
where $d$ is the distance from a body point to the obstacle's center, $\epsilon_d$ denotes a safety margin and is set as $\epsilon_d=4$\emph{cm}. Vanilla KMP and EKMP use the same parameters $\lambda=0.01$ and  
$k_h=4$. Other relevant parameters used by EKMP are $\beta=340$, $\lambda_{obs}=110$, $N=M=200$.

Evaluations are provided in Fig.~\ref{fig:letter:evolve},
showing that EKMP is capable of maintaining the shape of demonstrations and avoiding the obstacle while satisfying various constraints after very few iterations. The cost of obstacle avoidance is presented in Fig.~\ref{fig:letter:evolve}(\emph{f}), where fast decrease of obstacle avoidance cost is achieved via EKMP. 
Figure~\ref{fig:letter:pos:derivative} plots derivative of the predicted positions and the predicted velocities. We can see that the derivative relationship is indeed strictly ensured over the course of optimizing trajectories via EKMP.

\subsection{Evaluations on a Robotic Arm \label{subsec:robot:arm}}
Now, we employ EKMP to plan joint trajectories for a 7-DoF Kinova Gen3 robotic arm so that  reaching tasks can be accomplished while avoiding collisions of robot links with the obstacle and fulfilling various hard constraints.
We start with five demonstrations (comprising time and joint position) for a reaching task, as shown in Fig.~\ref{fig:joint:demos}, where the GMM modeling of demonstrations is also presented. 
We define two groups of constraints: \\
\textbf{(con-1)}: two desired points in terms of  joint position (7-D) and joint velocity (7-D), as depicted by the red circles in Fig.~\ref{fig:joint:eva}(\emph{a})--(\emph{c}) and (\emph{e})--(\emph{g});\\
\textbf{(con-2)}: 
\begin{equation}
\begin{aligned}
&f(\vec{q}_t)|_{t=6}=[0.60 \ -0.15 \ 0.60]^\trsp,\\
&f(\vec{q}_t)_z\geq0.4,
\end{aligned}
\label{equ:joint:con}
\end{equation}
where $f(\cdot)$ represents the forward kinematics of the robot arm and calculates Cartesian position of the robot's end-effector. $f(\vec{q}_t)_z\geq 0.4$ demands that the robot's end-effector should always stay above a horizontal plane with a height of 0.4\emph{m}. Note that $f(\cdot)$ is a nonlinear function of joint position and we here directly optimize joint trajectories.

The relevant parameters are
set as $\lambda=0.01$, $k_h=0.2$, 
$\beta=700$, $\lambda_{obs}=120$, the radius of the obstacle is $r=0.1$\emph{m}, $\epsilon_d=0.15$\emph{m}, $N=M=21$. Vanilla KMP also uses the same parameters. It is worth mentioning that trajectories with a length of 200 will be generated, which is far longer than $N$.  Since we only impose hard constraints over 21 points in order to reduce the computational cost in each iteration and thus some points from the generated trajectory may not strictly respect the constraints. However, this issue can be trivially solved by strengthening the constraints and in our evaluation the constraint $f(\vec{q}_t)_z\geq0.40+\epsilon_z$ with $\epsilon_z=2\times10^{-3}$ imposed over 21 points can ensure that $ f(\vec{q}_t)_z\geq0.40$ is strictly fulfilled over the generated trajectories (after  9 iterations) consisting of 200 datapoints.

Figure ~\ref{fig:joint:eva} depicts trajectory optimization within 10 iterations using EKMP. From Fig.~\ref{fig:joint:eva}(\emph{a})--(\emph{c}) and (\emph{e})--(\emph{g}), we can see that all optimized trajectories (green curves) satisfy the desired point constraints (marked by red circles) defined in \textbf{(con-1)}. Specifically, in Fig.~\ref{fig:joint:eva}(\emph{h}), the optimized trajectory at the last iteration goes through the desired Cartesian position (marked by `$\times$') and obeys the $z$-direction limit (depicted by the red plane), therefore respecting the constraints defined in \textbf{(con-2)}. Fast decrease of obstacle avoidance cost is also achieved,
as shown in Fig.~\ref{fig:joint:eva}(\emph{i}), where the cost is calculated using the generated joint trajectory (i.e., 200 datapoints).

For the sake of clear observation, Fig.~\ref{fig:joint:robot:eva} illustrates planned joint trajectories through executing them on the robotic arm. In contrast to vanilla KMP where robot links collide with the obstacle (Fig.~\ref{fig:joint:robot:eva}(\emph{a})),  joint trajectories generated by EKMP are capable of avoiding the obstacle under different hard constraints, see Fig.~\ref{fig:joint:robot:eva}(\emph{middle}  and \emph{right} graphs) where the joint trajectories correspond to the ones obtained after 10 iterations. Note that in our evaluations, under \textbf{(con-1)} the planned joint trajectories after 3 iterations can avoid the obstacle, while under both \textbf{(con-1)} and \textbf{(con-2)}, it only needs 1 iteration. 
This also provides an interesting insight, i.e., proper constraints in Cartesian space could be of help for fast planning of collision-free joint trajectories.

\section{Conclusions\label{sec:conclusion}}
We have extended KMP towards a more flexible framework capable of dealing with a bunch of key problems in imitation learning, including learning and adaptation, obstacle avoidance, linear and nonlinear hard constraints.  Several evaluations including 2-D writing tasks and joint trajectory planning for reaching tasks verified the performance of our framework. While other constraints such as stiffness and orientation are also important in many scenarios, it would be of interest to address them within 
the current framework.

\section*{Acknowledgement}
The author would like to thank Dr. Songyan Xin from the University of Edinburgh for the helpful discussion about evaluations on the robotic arm, and thank Dr. Jo\~{a}o Silv\'{e}rio from Idiap Research Institute for his comments on this paper.

\bibliographystyle{IEEEtran}
\bibliography{bibiography}

\end{document}